# Implementation of a ZED 2i Stereo Camera for High-Frequency Shoreline Change and Coastal Elevation Monitoring

José Pilartes-Congo[1,2], Matthew Kastl[1,2], Michael Starek[1,2], Marina Vicens-Miquel[1,2], Philippe Tissot[2]

[1] College of Engineering and Computer Science, Texas A&M University–Corpus Christi, Corpus Christi, TX, USA
[2] Conrad Blucher Institute for Surveying and Science, Corpus Christi, TX, USA

## Abstract

The increasing population, thus financial interests, in coastal areas have increased the need to monitor coastal elevation and shoreline change. Though several resources exist to obtain this information, they often lack the required temporal resolution for short-term monitoring (e.g., every hour). To address this issue, this study implements a low-cost ZED 2i stereo camera system and close-range photogrammetry to collect images for generating 3D point clouds, digital surface models (DSMs) of beach elevation, and georectified imagery at a localized scale and high temporal resolution. The main contributions of this study are (i) intrinsic camera calibration, (ii) georectification and registration of acquired imagery and point cloud, (iii) generation of the DSM of the beach elevation, and (iv) a comparison of derived products against those from uncrewed aircraft system structure-from-motion photogrammetry. Preliminary results show that despite its limitations, the ZED 2i can provide the desired mapping products at localized and high temporal scales. The system achieved a mean reprojection error of 0.20 px, a point cloud registration of 27 cm, a vertical error of 37.56 cm relative to ground truth, and georectification root mean square errors of 2.67 cm and 2.81 cm for $x$ and $y$.

## 1 INTRODUCTION

According to the National Oceanic and Atmospheric Administration (NOAA), as many as 127 million (roughly 40% of the U.S. population) lived in coastal counties in 2014 [1]. Therefore, continuous monitoring of changes in the shoreline, coastal elevation, and inundation helps to preserve the well-being of coastal residents and financial interests. While conventional land surveying methods and airborne data can be used to observe such changes, it would require far more labor and financial resources for high-frequency observations (e.g., hourly or daily). Hence, the objective of this study is to address these issues by implementing a low-cost ZED 2i stereo camera. The camera is mounted at Horace Caldwell Pier (Port Aransas, Texas) and is used to generate 3D point clouds, digital surface models (DSMs) of beach elevation, and georectified images. These can then be used to track changes in shoreline and local elevation over time. This low-cost system requires no field personnel on-site, which is a benefit that conventional surveying techniques lack. In addition, the ZED 2i-derived products are compared to those generated via uncrewed aircraft system (UAS) based structure-from-motion (SfM) photogrammetry, using a DJI Phantom 4 RTK UAS platform. Both datasets are supported by ground truth from real-time kinematic (RTK) global navigation satellite systems (GNSS) by connecting to the Texas Department of Transportation real-time network. The derived mapping products are referenced to the North American Datum of 1983 (epoch of 2011) and the North Vertical Datum of 1988.

Intrinsic camera calibration is used to obtain the parameters that determine the internal geometry of a camera, such as focal length and lens distortion [2]. Appropriate calibration helps to ensure that camera-generated mapping products are reliable. It is recommended that the images used for intrinsic calibration are captured from different angles and distances [3]. Ideally, a camera would undergo intrinsic calibration in a controlled environment before its deployment in the field.

The use of cameras for coastal monitoring and perspective mapping has been common in recent years. SurfRCaT, a python-based tool, was proposed in [4] to calibrate and rectify images from existing cameras across U.S. coastal sites. SurfRCaT is effective for continuous monitoring but does not consider lens distortion during calibration. In their case study, the authors reported errors ranging from 3.8 m to 32.7 m, with the highest errors observed on points farthest from the camera [4]. In [5], the Coastal Imaging Research Network (CIRN) tool is introduced as a series of MATLAB scripts used to georectify oblique imagery for creating coastal mapping products. A case study was discussed that resulted in horizontal errors of 14 cm [5]. In [6], a PhenoCam camera was used for ecosystem-level studies, and the authors were able to reach 10 cm accuracy at distances within 100 m. While the methods and resulting accuracies obtained in these studies have varied, they all suggest degrading oblique mapping accuracies as the distance from the camera increases. This is consistent with statements in [7] and happens largely because of scale variations across the image space. The methods explored in this study can aid decision-making in coastal environments and promote the use of affordable means for coastal monitoring.

## 2  STUDY SITE

This study took place at Horace Caldwell Pier - Port Aransas Beach, Texas (Fig. 1). The area of interest comprises approximately 45 m cross-shore and 60 m alongshore and is a pedestrian section of the beach characterized by fine beach sand and relatively flat terrain. This location was chosen for being the closest gulf-facing pier to the base of operations that was fully operational at the start of this project.

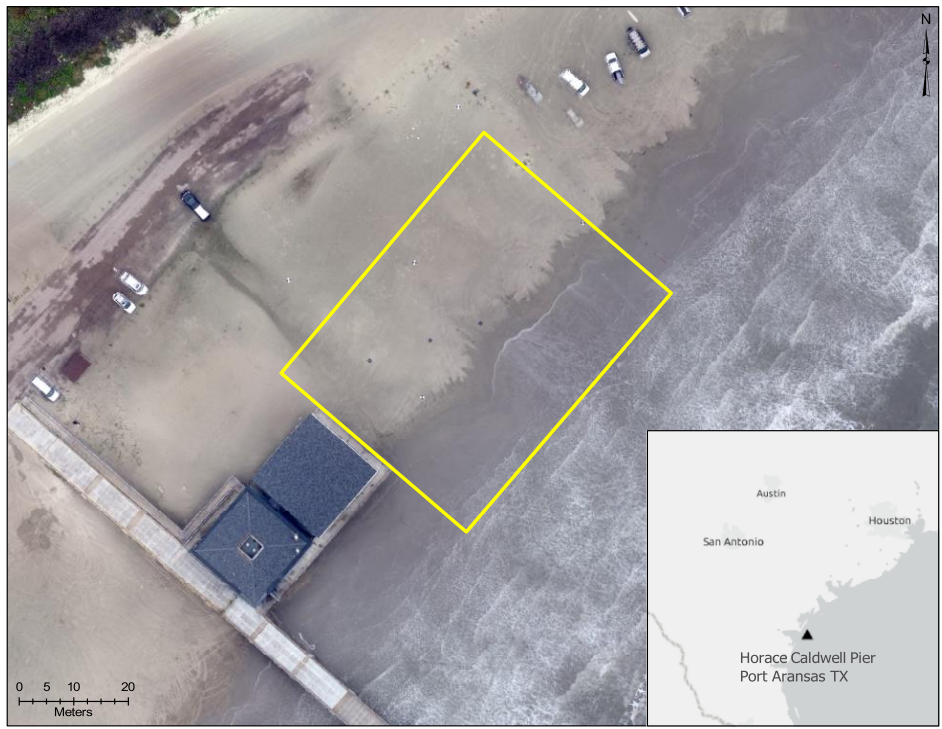

Figure 1: Area of interest at Horace Caldwell Pier.

## 3  METHODOLOGY

### 3.1  Hardware and Software

The system used in this study consists of a ZED 2i stereo camera ($500) manufactured by Stereolabs and an NVIDIA Jetson Xavier-NX computer ($1,400) [8]. The ZED 2i has two 4 MP camera sensors with f/1.8 aperture, a baseline separation between the cameras of 12 cm, and a 160º field-of-view angle [8, 9]. This project uses a camera with a 2.1 mm focal length. Yet, a 4 mm option is also available [8]. The computer has a 348-core NVIDIA Volta GPU (48 tensor cores), a 6-core NVIDIA Carmel ARM 64-bit CPU, a 256GB M.2 NVMe SSD, and it operates in temperatures ranging from -20º C to 60º C [10]. The external connection of the system is handled by an Airlink RV55 LTE cellular modem of 4G LTE capability and peak downloading and uploading speeds of 150 mbps and 50 mbps, respectively [11]. Although this project uses Python to capture imagery in 30-minute intervals, the ZED 2i system is also compatible with C++, C#, and C programming languages [12]. All acquired data is compressed and uploaded to an Amazon S3 cloud storage for safekeeping. In addition, a DJI Phantom 4 RTK UAS was used for comparative assessment. This platform uses a 20 MP FC6310R digital camera with a focal length of 8.8 mm [13]. The standard build for this UAS, without premium features, is worth roughly $10,000. In terms of software, MATLAB was used for intrinsic camera calibration and QGIS for image georectification. CloudCompare was used to register the 3D point cloud from the ZED 2i. Pix4Dmapper was used to process the data collected using the DJI Phantom 4 RTK. Final maps were created using ArcGIS Pro. All data processing was performed on a 64-bit Windows 10 computer with the following specifications: Intel(R) Core(TM) i9-7940X CPU @ 3.10 GHz, 128 GB RAM.

### 3.2 Processing Workflow and Accuracy Assessment

The first step in this experiment performed an intrinsic camera calibration using the Stereo Camera Calibrator Tool in MATLAB and the checkerboard calibration method (see Fig. 2). Each of the stereo cameras captured a series of 40 images of the checkerboard from different distances and angles, leading to a total of 80 photos. The goal of this step was to retrieve the internal parameters of the camera, namely focal length, principal center, and radial and tangential distortion coefficients.

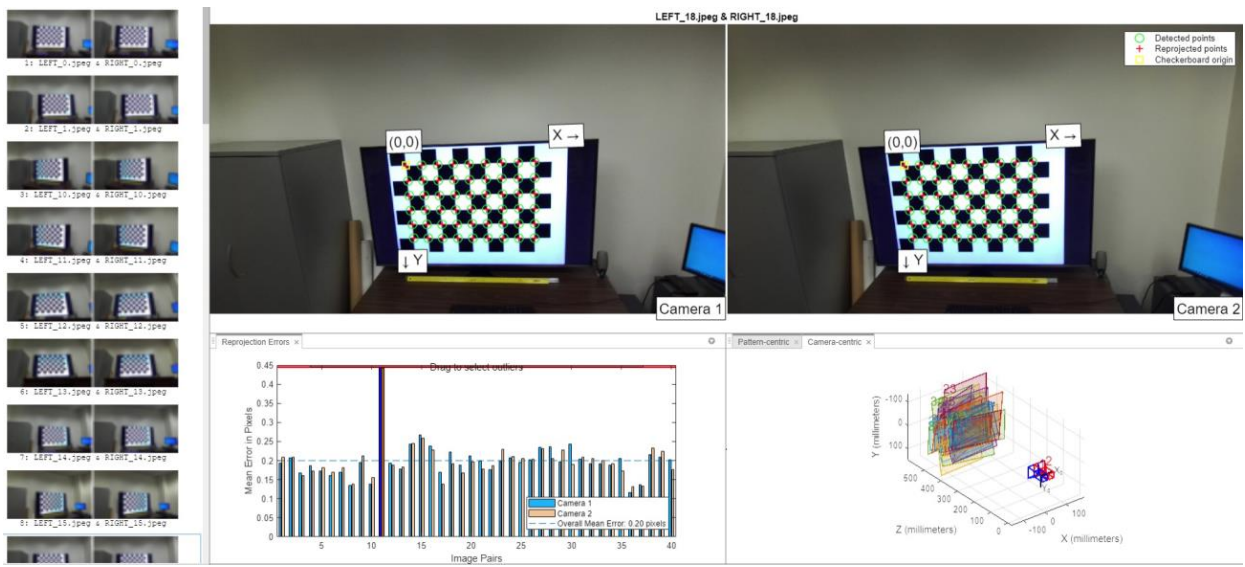

Figure 2: Interface of MATLAB during intrinsic calibration.

Although Stereolabs performs extensive factory calibration on its ZED 2i cameras [9, 14], periodic re-calibration helps to ensure the best accuracy for resulting data products. Using the API provided in [14], the initial intrinsic parameters (in pixels) resulting from the factory calibration were as follows: focal length ($fx$, $fy$) values of 1060.70 and 1060.70; principal center ($cx$, $cy$) values of 950.42 and 572.89; and values of 0 across all radial ($k1$, $k2$, $k3$) and tangential ($p1$, $p2$) distortion coefficients. These were then compared against the results obtained using the Stereo Camera Calibrator Tool.

After the intrinsic calibration, the camera was mounted on the pier to oversee and capture images of the area of interest (see example in Fig. 3). In addition, a series of GCPs were distributed across the camera's field of view to aid the image georectification process. The GCPs were surveyed via RTK GNSS using a 5-second average at a 1 hz sampling rate. In this experiment, georectification was performed on the right camera photo using a projective transformation in QGIS and a cubic resampling method aided by the GCPs. The 3D point cloud was generated using the manufacturer's depth sensing API and was colorized using R, G, B, and A channels. Next, the point cloud was registered using the alignment tool in CloudCompare and the GCPs. In turn, the registered point cloud was used within the rapidlasso/LAStools for vertical accuracy assessment (using the *lascontrol* module) and to generate the desired DSMs (using the *las2dem* module). Lastly, a shapefile was created to define to clip the final DSM. This was necessary to remove unwanted elements and noisy features caused by the reduced range of the camera. When using Pix4Dmapper to process the UAS data, the standard 3D mapping preset with its default parameters was chosen.

With respect to the accuracy assessment, the intrinsic calibration error is evaluated based on error metrics provided by MATLAB. The georectification accuracy assessment was performed by means of the root mean square error (RMSE), which is a statistical metric used to quantify the closeness between observed and expected values. Here, the observed values are tied to the georectified image, and the expected values are provided by the GCPs. As mentioned earlier, only the right image is used in this part of the experiment. Thus, the RMSEs presented herein appraise the registration accuracy of the georectification process and they are calculated using equation 1, where $n$ represents the number of data points in a sample; $x_i$ and $\hat{x}_i$ are the observed and expected $x$ values for a given point $i$; and $y_i$ and $\hat{y}_i$ are the observed and expected $y$ values for the same point $i$. Implementing the RMSE metric for photogrammetric purposes assumes that systematic errors have been addressed accordingly [15]. Performance metrics of the UAS were provided in the processing report generated after processing with Pix4Dmapper.

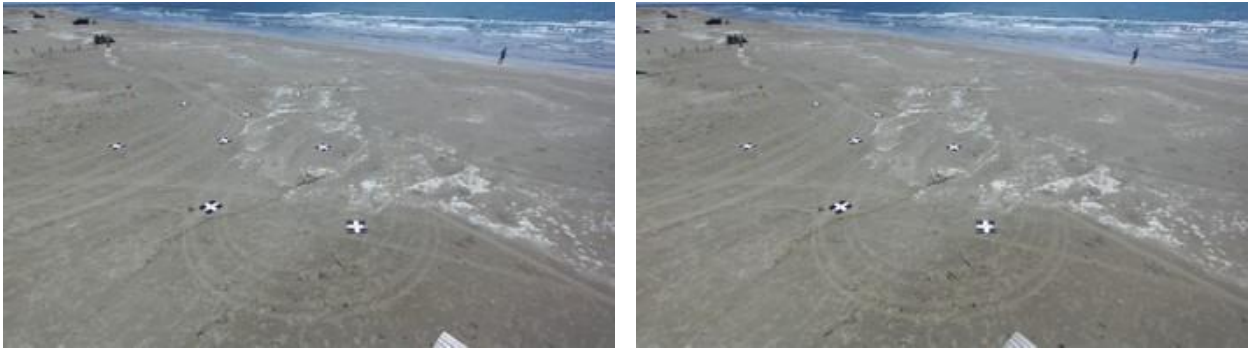

| (a) ZED (left photo) | (b) ZED (right photo) |

Figure 3: ZED 2i stereo camera left and right photos.

$$RMSE_x = \sqrt{\sum_{i=1}^{n} \frac{(x_i - \hat{x}_i)^2}{n}}; RMSE_y = \sqrt{\sum_{i=1}^{n} \frac{(y_i - \hat{y}_i)^2}{n}} \quad (1)$$

## 4 RESULTS AND DISCUSSION

Intrinsic camera calibration yielded a mean reprojection error of 0.20 px. In terms of intrinsic camera parameters, minimal differences were observed between the values provided by Stereolabs and those retrieved using the Stereo Camera Calibrator Tool in MATLAB as shown in Table 1. Observe that some level of distortion ($k$ and $p$ values) is still observed despite the manufacturer's extensive calibration, hence the importance of period camera calibration for reliable mapping.

Table 1: Intrinsic camera calibration parameters (Stereolabs vs MATLAB Calibrator Tool).

|  | Stereolabs | MATLAB |
|---|---|---|
| $fx$, $fy$ | 1060.70, 1060.70 | 1059.85, 1060.14 |
| $cx$, $cy$ | 950.42, 572.89 | 951.58, 573.87 |
| $k1$, $k2$, $k3$ | 0, 0, 0 | 0.0046, -0.0715, 0.1904 |
| $p1$, $p2$ | 0, 0 | -0.0003, -0.0019 |

After georectification, the $RMSE_x$ and $RMSE_y$ relative to the network of seven GCPs were 2.67 cm and 2.81 cm, respectively. Like similar studies (e.g., [4, 6]), errors of slightly higher magnitude were observed as the distance from the camera increased. In contrast, the $RMSE_x$ and $RMSE_y$ values for the UAS were 0.88 cm and 0.91 cm, respectively. Despite these differences, there was a positive correspondence between the ZED 2i georectified image and UAS orthomosaic in the area of interest (see Fig. 4a). Also, when using the ZED 2i dataset, a point cloud registration accuracy of 27 cm was obtained, and *lascontrol* yielded a vertical error of 37.56 cm between ground truth (GCPs) and the point cloud. Fig. 4b shows a DSM generated from the ZED 2i point cloud. The red area on the west end was caused by the edge of the pier that was visible from the photos. The fact that the model is able to detect the higher elevation shows how precise it can be when mapping in close proximity. The red area on the northeast side of the map is hypothesized to be an artifact caused by the limited range of the camera. The UAS point cloud required no registration, and the vertical error provided by Pix4Dmapper's report was 0.66 cm.

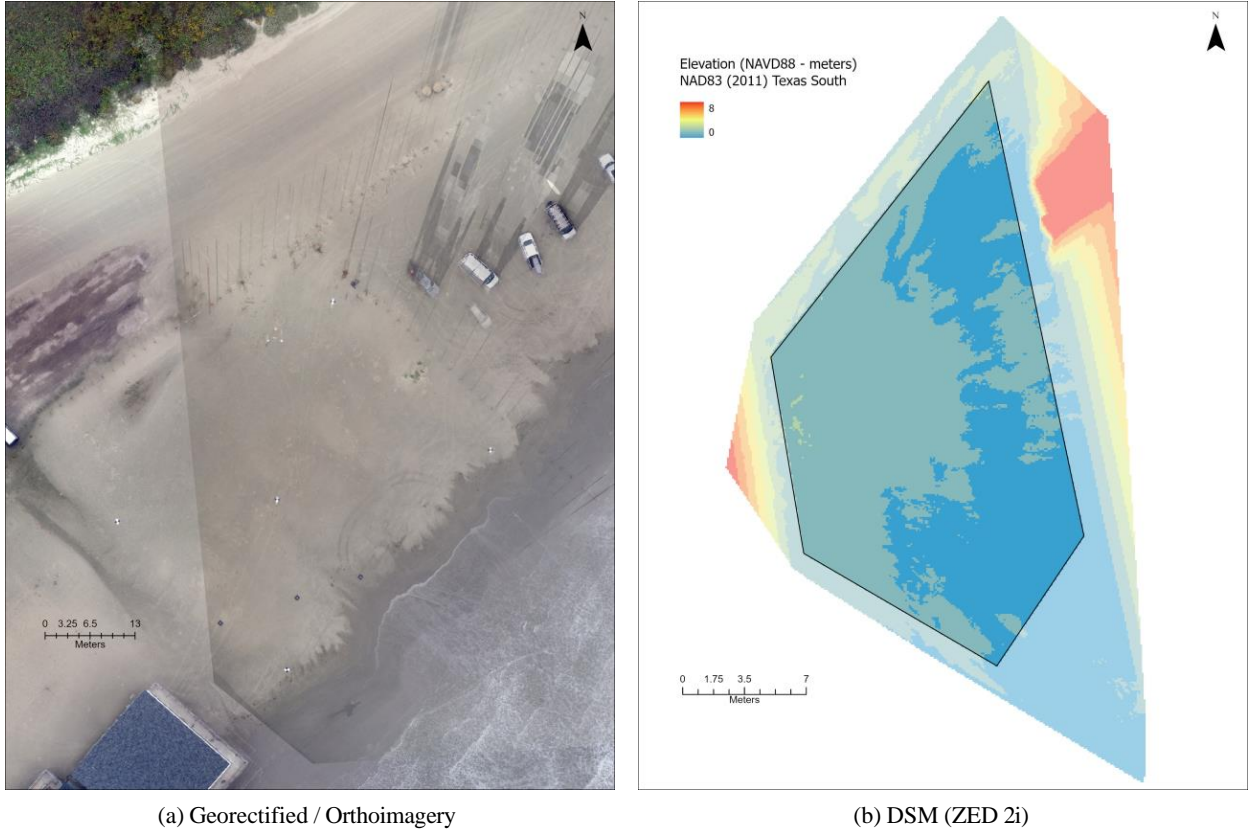

(a) Georectified / Orthoimagery  (b) DSM (ZED 2i)

Figure 4: (a) ZED 2i georectified image (darker shade) overlaid on the UAS-generated orthomosaic (lighter shade); (b) DSM generated from the ZED 2i registered 3D point cloud.

## 5  CONCLUSION

This work implements a low-cost ZED 2i stereo camera to collect imagery that is used to create 3D point clouds, DSMs, and georectified imagery to monitor coastal dynamics. The major benefit of this system is that it helps to create a time series of data at high temporal resolution, which is a task that is not always possible with conventional surveying methods. The ZED 2i system can provide georectified imagery within centimeter-level horizontal errors but it is plagued by limitations such as short range and reduced focal length that must be considered before use. Experiments with the DJI Phantom 4 RTK provided higher levels of accuracy (within the millimeter range) than the ZED 2i. The next steps in this research consist of (i) improving the ZED 2i obtainable vertical accuracies and (ii) automatically detecting the wet/dry shoreline position and elevation using the work proposed in [16]. The shoreline information will be used to train an artificial intelligence model with the goal of predicting coastal inundation.

## 6  ACKNOWLEDGEMENTS

Thanks to J. Millien, J. Berryhill, K. Colburn, S. Stephenson, and the CBI operations team for their help with camera installation and field acquisition efforts. This work was supported in part by the NSF AI2ES, and in part by the National Science Foundation (NSF) under Award 2112631. Any opinions, findings, conclusions, or recommendations expressed herein are those of the authors and do not necessarily reflect the views of the funding agencies.